\documentclass[sigconf]{acmart} 
\AtBeginDocument{%
  \providecommand\BibTeX{{%
    \normalfont B\kern-0.5em{\scshape i\kern-0.25em b}\kern-0.8em\TeX}}}

\usepackage{tabularx}
\usepackage{pifont}

\copyrightyear{2023} 
\acmYear{2023} 
\setcopyright{rightsretained} 
\acmConference[CHI '23]{Proceedings of the 2023 CHI Conference on Human Factors in Computing Systems}{April 23--28, 2023}{Hamburg, Germany}
\acmBooktitle{Proceedings of the 2023 CHI Conference on Human Factors in Computing Systems (CHI '23), April 23--28, 2023, Hamburg, Germany}\acmDOI{10.1145/3544548.3581483}
\acmISBN{978-1-4503-9421-5/23/04}

\begin{document}


\title{Amortised Experimental Design and Parameter Estimation for User Models of Pointing}

\author{Antti Keurulainen}
\email{antti.keurulainen@bitville.com}
\affiliation{%
  \institution{Aalto University}
  \city{Espoo}
  \country{Finland}
}

\affiliation{%
  \institution{Bitville Oy}
  \city{Espoo}
  \country{Finland}
}

\author{Isak Westerlund}
\email{isak.westerlund@bitville.com}
\affiliation{%
  \institution{Bitville Oy}
  \city{Espoo}
  \country{Finland}}

\author{Oskar Keurulainen}
\email{oskar.keurulainen@bitville.com}
  \affiliation{%
  \institution{Bitville Oy}
  \city{Espoo}
  \country{Finland}}

\author{Andrew Howes}
\email{a.howes@bham.ac.uk}
\affiliation{%
\institution{University of Birmingham}
\city{Birmingham}
\country{United Kingdom}}
\affiliation{%
\institution{Aalto University}
\city{Espoo}
\country{Finland}}

\renewcommand{\shortauthors}{Keurulainen et al.}

\begin{abstract}
User models play an important role in interaction design, supporting automation of interaction design choices. In order to do so, model parameters must be estimated from user data. While very large amounts of user data are sometimes required, recent research  has shown how experiments can be designed so as to gather data and infer parameters as efficiently as possible, thereby minimising the data requirement. In the current article, we investigate a variant of these methods that amortises the computational cost of designing experiments by training a policy for choosing experimental designs with simulated participants. Our solution learns which experiments provide the most useful data for parameter estimation by interacting with in-silico agents sampled from the model space thereby using synthetic data rather than vast amounts of human data. The approach is demonstrated for three progressively complex models of pointing.
\end{abstract}



\begin{teaserfigure}
  \includegraphics[width=1.0\textwidth]{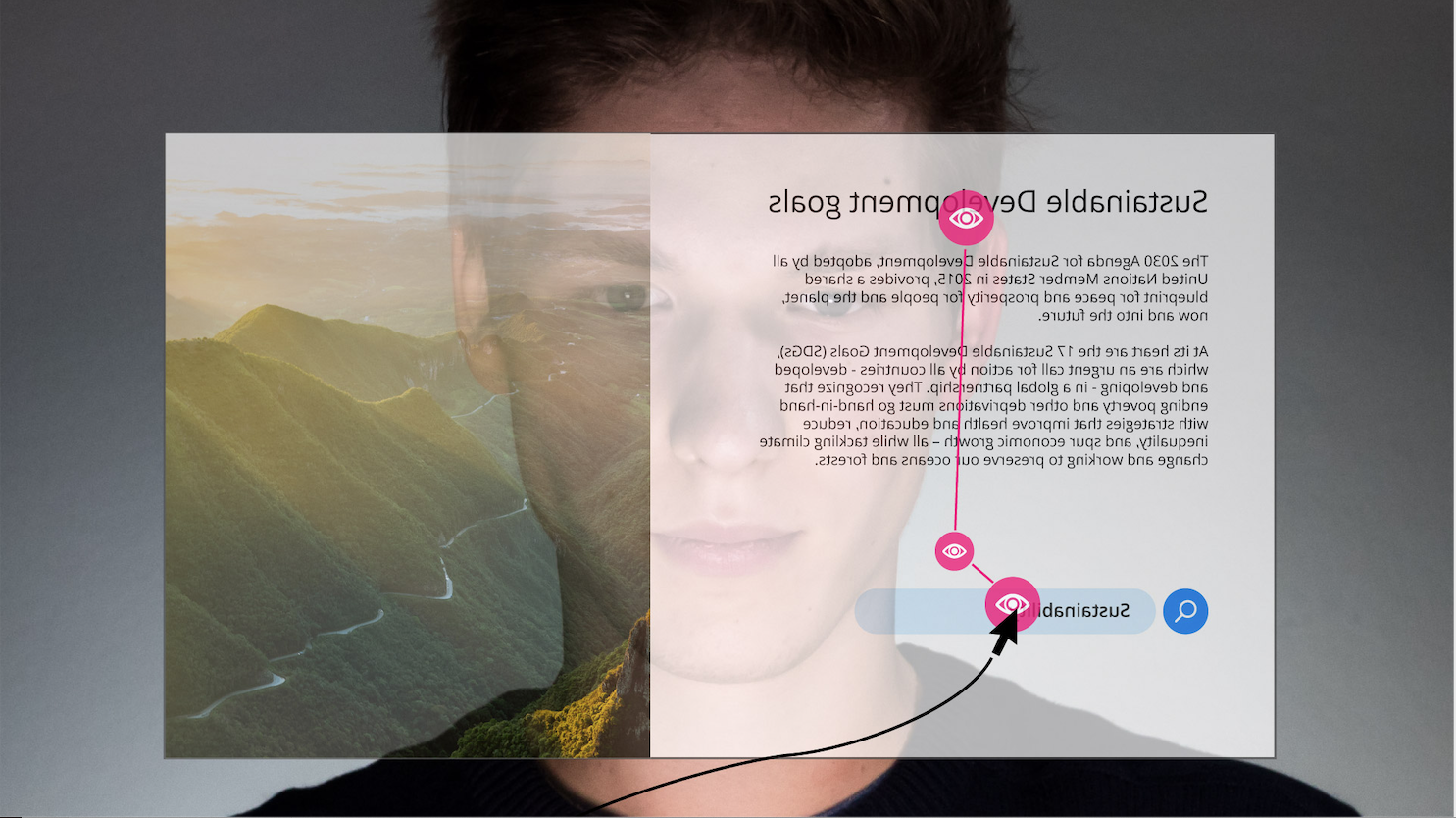}
  \caption{An intelligent interactive system views a user through sensors that include a keyboard and mouse and sometimes, camera, microphone and/or eye-tracker. What is the best way to infer a model of the user from these data? In the figure gaze path is represented in red, mouse movements in black and text entered in blue. A camera captures the user's gaze direction.}
  \Description{A computer views a user. The figure illustrates how an intelligent interactive system views a user through sensors that include a keyboard and mouse and sometimes, camera, microphone and/or eye-tracker.}
  \label{fig:teaser}
\end{teaserfigure}


\maketitle

\section{Introduction}
User models take many forms in HCI, from simple lists of `psychological factors' including, perhaps, personality variables and/or product preferences to cognitive models that simulate the processing of information in the mind. 
Some of the latter have focused on constraints imposed by the human perceptual and motor systems, others on the structure of human memory and others on control (what to do next).
While these user models provide one of the theoretical anchors for the discipline, they are difficult to construct and difficult to fit to behaviour -- sometimes requiring vast amounts of data from an expanding range of sensors (Figure \ref{fig:teaser}). In the current article, we explore one particular approach to addressing  this latter problem by automating model parameter estimation for three pointing tasks. The approach involves the design of an optimal sequence of experimental trials (or designs) that maximise the relevance of the available information.

In the statistics literature `optimal experimental design' (OED) is the problem of choosing which experimental trial to do next so as to maximize some objective. For example, an HCI researcher may want to measure the effect of a new pointing device on a user's movement accuracy, but how far away and how big should the movement target be on each successive trial so as to maximize the information gained from observing the user performing the task? We present an approach to efficiently solving this problem for estimating user model parameters in HCI.

We argue that the presented approach has the potential to enhance the contribution that user models make to HCI by providing interactive systems with the means to automatically and rapidly fit user models to individual users and thereby personalise interaction so as to best fit the requirements of the individual. This capability is also important to cooperative/collaborative Artificial Intelligence (AI), that is the problem of how to get machines to work with people. Personalisation and collaboration are important objectives for HCI, in part, because they directly address the desire to design interaction for diverse users. While we do not investigate personalisation per se in the current article, we believe that user modeling is crucial to the future of personalisation and that this potential will only be fulfilled if the parameter estimation problem can be solved.

Fitting a user model to an individual was difficult in the early days of cognitive modeling, in part because models such as GOMS \cite{Card1983}, were constructed manually. Production rules that mapped goals into actions were written by an analyst. Model constructing consisted of a painstaking and iterative process of protocol analysis, and production rule writing.

Since then significant advances have been made on automatic construction of models. Rather than hand-coding production rules, modern modelling techniques now automatically learn a \emph{control policy} (task knowledge) using deep reinforcement learning. In particular,  machine learning can be used to learn a model's control policy through exploration of simulated interaction.\footnote{The control policy is a function that maps observations into actions. It can be learned with a number of machine learning algorithms.} 
This approach has  been successfully applied to the automatic construction of user models of menu search, decision making, gaze-based interaction, hierarchical control, and touchscreen typing \cite{chen2021adaptive, jokinen2021touchscreen, chen2017cognitive, chen2021apparently, Chen2015emergence, gebhardt2020hierarchical, do2021simulation, oulasvirta2022computational}. However, fitting the quantitative parameters of these models to individual humans still requires a significant contribution from the analyst. 

One of the difficulties with fitting learning-based user models to human data is that the control policy must be trained to make predictions for each set of possible parameters. For example, in the gaze-based interaction model reported in \cite{chen2021adaptive}  oculomotor noise and perceptual noise parameters are properties of an individual user. Both parameters introduce uncertainty in aimed movements to a target. The optimal control policy chooses an aim  point for  a target in accordance with these uncertainties. For example, when oculomotor noise is high then it makes sense to deliberately undershoot the target and then make a corrective submovement. This is because on average this results in a lower overall movement time than overshooting (which takes longer) and correcting. 

Unfortunately then, while automatically learning human-like control policies solves part of the  user modelling problem, it does not solve the parameter estimation problem. User models typically have many parameters and the control policy, and therefore behaviour, is adapted to these parameter values. For example, it is known that the human control policy for pointing is adapted to noise both in the motor system and the visual system. In general, the parameter estimation problem is to find a set of values for model parameters such that the predicted behaviour of an individual user is as close as possible to the observed (measured) behaviour. It is quite often formalised as an optimisation problem; how to generate the best estimate of the parameters from the available data. Typically, the objective will be to find parameters that minimise the difference between the model behaviour and human behaviour, that is maximising fit.

Estimating parameters so as minimise the discrepancy between model and human requires high quality data; a fact that, as we have said, has given rise to work on  optimal experimental design (OED) \cite{myung2013tutorial}. This problem has been conceptualised as a problem of how to maximise expected information gain (EIG). In other words, the purpose of an experiment is to maximise the information that is gained about the parameter values that best fit the model to the human. Various approaches have been proposed to this problem. In one class of approaches, Bayesian Optimal Experimental Design (BOED), the idea is to choose experiments that maximise EIG between the prior probabilities of all possible parameters values and a posterior distribution that is conditioned on the expected observations \cite{ryan2016review}. BOED does not only give a point estimate of the best fitting parameter values but also a posterior probability that these (and all other parameter values) are the best fit. 

In some approaches, the choice of experiment is `amortised' meaning that a near-optimal policy for choosing experiments is computed before the deployment of the policy for parameter estimation. In HCI, the advantage of amortisation is that design of experiments for estimating user models can be computed without slowing interaction with the user. Another advantage of amortisation is that experimental design can be non-myopic. This means that, rather than maximising EIG for each individual experiment, instead it can be maximised for a whole sequence of experiments. One approach to amortisation involves defining the optimal experimental design problem as a reinforcement learning problem.

\begin{figure*}
  \includegraphics[width=0.8\textwidth]{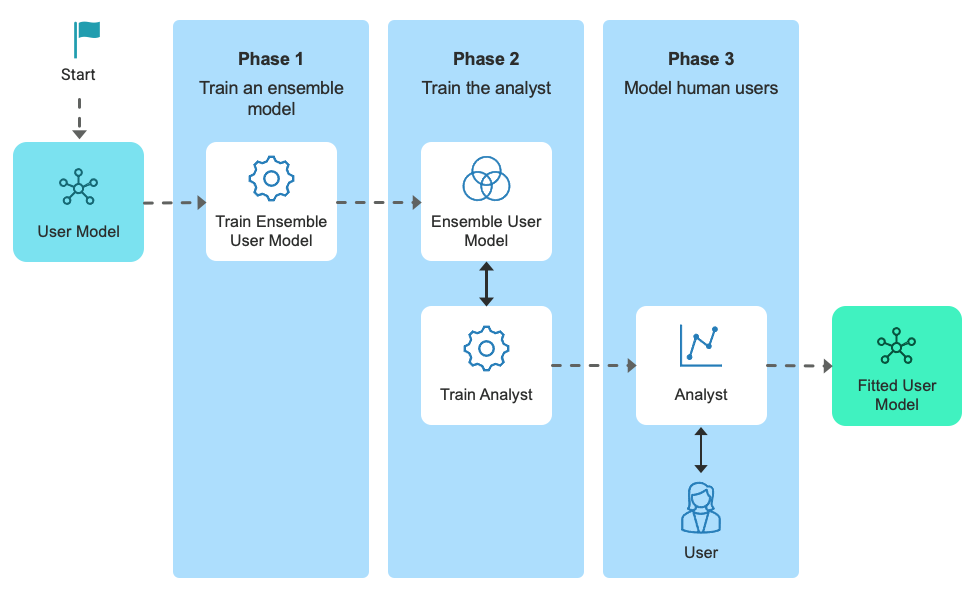}
  \caption{Our approach takes a user model as input. This user model has prior distributions over parameters $\theta$ but has not been fitted to individual human behaviour. In Phase 1, an Ensemble Cognitive Model (ECM) is trained to perform the task for the distribution of possible parameter values. In Phase 2, the cognitive model Analyst is trained to conduct the best sequence of experiments for determining the model parameters. In Phase 3, the trained Analyst is deployed with users and a fitted model it generated as output.}
  \Description{An overview figure of the approach illustrating the three phases. The three phases are training the ensemble model, training the analyst and modelling the human user. A more detailed explanation of the phases is given in section 1}
  \label{fig:overview}
\end{figure*}

In the current article we propose a new approach to user model parameter estimation in HCI that takes advantage of the recent advances described above. The contribution is in offering a novel and practical method for estimating user model parameters through amortising the cost of choosing experiments. While the proposal is for a general method, we demonstrate its viability in this paper for pointing tasks. An overview of the approach is illustrated in Figure \ref{fig:overview}. The approach estimates the parameters of a user model for an individual human user (Phase 3), having  previously computed an `ensemble' user model for all possible parameter combinations (Phase 1) and then an optimal sequential experimental design policy (Phase 2). 
In phase 1, the ensemble model of the space of possible users is trained to perform the task for the distribution of possible parameter values and the distribution of possible task environments. The ensemble approach is an important recent advance in user modelling that is described in more detail below \cite{kwon2020inverse, moon2022speeding}.
In phase 2, the Analyst is trained to conduct the best sequence of experiments for determining the model parameters. Simulated users are randomly sampled given the parameter distribution and the Analyst learns to fit the model parameters to the simulated user. In phase 3, the trained Analyst conducts experiments on users and generates parameter fits.

Phase 1 amortises the cost of computing the implications of different parameter values for model behaviour. Phase 2 amortises the cost of computing a non-myopic sequential policy for choosing experiments and Phase 3 takes advantage of the computation conducted in Phases 1 and 2 in order to optimally gather data and estimate user model parameters without any user-noticeable interaction latency. 

One confusion that arises about our approach is that once we have trained a simulator model for all possible parameter values (the 'ensemble' user model) then there is  nothing else to learn. The confusion is resolved once it is realised that the 'ensemble' is a model of the distribution of all possible users (technically, a distribution over the parameters of the user model) and all possible tasks within the defined space. The assumption is that if we know the exact parameter values for a specific real-world user, then this model will accurately describe their behaviour, but the challenge is that, given a real-world user, we do not know the best parameters. We need the Analyst to design experiments and gather the data from the user in order to find the best parameters in the ensemble model.

For training the Analyst, we generated hypothetical users by sampling parameters from the ensemble distribution, and exposed these known parameter values to the Analyst as a training signal, but there was no expectation that any individual real-world user will have the same parameter values. At test time, the Analyst had no knowledge of the user parameters (and the sets of training-time and test-time users are disjoint), and these parameters were inferred from the designed experiments.

Another confusion is that it might seem that there is no point in learning about an individual user when Analyst already has a model of the  distribution of  all possible users. But, this model can only simulate an  \emph{specific} user if it knows the correct parameters for that user. The point of Analyst is to infer the parameters of a user with unknown parameters (e.g. a new human user whose behaviour has not been observed before). Analyst needs to infer the best values of those parameters, and only then can the user model simulate that particular user. Further, it is the capacity to simulate a specific user that promises a role for Analyst in interaction personalisation and cooperative Artificial Intelligence.

In what follows, we review the existing literature, formally define our approach, test it on two abstract tasks chosen so as to demonstrate the generality of the approach, and then report three studies of parameter estimation for user models of pointing. In the first of the studies, we demonstrate the effectiveness of the approach for estimating the parameters of a pointing user model from mouse click data. The user model has a single parameter for movement noise that gives rise to Fitts's Law like behaviour through a learned policy that generates multiple submovements to achieve point and click goals.

In the second study, we extend the approach to simulated eye-movement data for gaze-based pointing. These data consist of variable length sequences of interleaved saccades and fixations. In this gaze-based model there are two parameters, one for oculomotor noise and the other for perceptual noise. This gives rise to a potential identifiability problem. 

In the third study, we explore the capacity of the approach to not only identify perceptual/motor noise parameters but to also determine user preferences. Here the gaze-based model is applied to an interface in which pointing is achieved by looking at targets and pressing a button (a key on the keyboard). Because an experimental trial can be terminated at any time by pressing the button it gives rise to a speed/accuracy trade-off and the user's policy is optimised for their preference, or otherwise, for accuracy over speed. In this study both performance time and errors are used in the estimation of model parameters.

\section{Background}

\subsection{User models}

User models -- computable representations of psychological constraints on interaction -- have been influential in HCI since its inception. GOMS, a rule-based model for representing hierarchical task knowledge, provides a formalism for HCI researchers to conduct detailed task analyses \cite{Card1983}. The Model Human Processor (MHP), a theory of the temporal properties of cognitive resources, supported the prediction of task performance time. ACT-R and EPIC provide means to simulate cognition and action; ACT-R with a particular focus on human memory and EPIC on constraints imposed by perceptual/motor systems \cite{Anderson2004, Kieras2014, Zhang2014}. Fitts's Law, a mathematical formulation of the relationship between task difficulty and movement time, became a particularly influential model of pointing \cite{Mackenzie1992, zhai2004speed}. Computationally rational models, based on machine learning problems but with cognitive bounds, provided a means to automatically learn control policies. Rather than hand crafted production rules, computationally rational models derive predictions by learning a control policy that is bounded only by human like resource constraints \cite{oulasvirta2022computational, lewis2014computational, howes2009rational}.

Most of the approaches to user modelling described above remain actively and productively investigated but our focus here is on computationally rational models \cite{oulasvirta2022computational}. While it has a distinctive approach to the automation of a control policy, it shares a need for new approaches to parameter estimation. In what follows we look in detail at three computationally rational models in order to further understand the parameter estimation problem.

One approach to computational rationality involves defining an interactive cognitive task as a Markovian problem and solving it using machine learning, usually reinforcement learning. For example, multi-attribute decision making can be defined as a Partially Observable Markov Decision Problem (POMDP) in which information about relevant attributes is gathered using saccadic eye movements. The reward function specifies a trade-off between speed and accuracy such that information is only gathered if the benefits to decision accuracy outweigh the temporal cost \cite{chen2017cognitive}. As a consequence, once an optimal control policy has been learned, it generates attribute-wise, rather than option-wise information gathering -- much like humans. While automatically acquired optimal control policies address a major part of the user modelling problem, they leave open the question of how to set model parameters. In the case of multi-attribute decision making the predictions are only as good as the attribute weights that define the user preference function.

Similarly, gaze-based interaction can be defined as a POMDP in which partial observations are constrained by foveated vision and saccadic eye-movements by oculomotor noise \cite{chen2021adaptive}. The model of foveated vision imposes increasing localisation error with eccentricity of the target from the fovea. The solution to this POMDP is an optimal control policy that -- again like humans -- undershoots the target in order to minimise movement time (because undershoots take less time than overshoots). Again, the capacity to automatically generate a control policy is an important advance but it leaves open the question of how the parameter values are set so as to model individual users. In the case of gaze-based interaction parameters include the perceptual and oculomotor noise weights as well as the saccade duration intercept and slope parameters.

An important aspect of user modeling is determining the prior distribution of possible parameter values. This distribution can be thought of as an hypothesis space that covers all of the possible behaviours of the population of users. It is constructed using knowledge that is available before performing experiments on individual humans. This prior parameterisation of the model induces an ensemble of possible  user models, each of which can be expressed in the form of a particular POMDP. Importantly, this modelling framework is agnostic to the amount of prior information available, since the experimenter can specify arbitrary prior distributions for the parameters of the model. In the case of limited available prior knowledge, the simulator implementing the model can be initialised with non-informative priors, thus describing a diverse distribution of users with highly varying cognitive bounds and preferences. Once the prior distribution of possible user models is established then experiments on humans can be used to determine which user model (i.e. which parameter settings) best fits.

\subsection{Adaptive Experimental Design in ML} 

In Bayesian approaches to experimental design the starting assumption is that the posterior probability of a parameter value given an experiment is proportional to the likelihood of the data times the prior $p(\theta | y,d) \propto p(y|\theta,d) \times p(\theta)$. The key question is how to choose experiments that maximise the utility of the data. In Bayesian Optimal Experimental Design (BOED) it is assumed that the objective is to choose experiments that maximise Expected Information Gain (EIG). BOED usually relies on  a likelihood model $p(y|\theta,d)$ for predicting the probability of data $y$ (the outcome of an experiment) given an experimental design $d$ and parameter values $\theta$. The objective is then to optimise EIG. EIG can be thought of as the \emph{mutual information} between $\theta$ and $y$.

BOED methods have been successfully applied to the design optimisation problem in various settings \cite{chaloner1995bayesian,atkinson2007optimum}. However,  BOED  can require computationally expensive calculations, such as updating the posterior or estimating the mutual information between the model parameters and experiment outcomes. As these calculations are needed between the time steps of the experiment, this approach becomes impractical for many real-life settings. 
More recent work has amortised the cost of experiment selection using pre-trained deep neural networks. As an example, Foster et. al \cite{foster2021deep} suggest a policy network, parameterised by a deep neural network, to produce informative experimental design values. In their approach, the loss function is based on calculating a lower bound of the mutual information instead of costly exact values. This work was extended in \cite{ivanova2021implicit}, in which likelihood functions can be unknown, thus expanding this approach to implicit models. Another line of work is based on using the mutual information as the main criteria for selecting design values, but using neural estimators on the mutual information or its lower bounds \cite{belghazi2018mutual,kleinegesse2020bayesian,kleinegesse2021gradient,poole2019variational}. Blau et al \cite{blau2022optimizing} present a Reinforcement Learning formulation for design optimisation. They defined sequential experimental design as a Markov decision process (MDP), highlighting the strong exploration capability of RL-based methods. In their approach, the optimisation target is the lower bound of expected information gain (EIG), comparable to the DAD method in \cite{foster2021deep}.

\section{Theory} 

We present the theory in two parts. In the first part, we describe a user model with parameters that must be estimated from data. The user model is an example of a class of simulation-based reinforcement learning model that has recently become influential in HCI \cite{Chen2015emergence, chen2017cognitive, jokinen2021touchscreen, oulasvirta2022computational} but which because of their complexity currently lack adequate parameter estimation methods (though see \cite{kangasraasio2018inverse, moon2022speeding}.

In the second part of the current section we introduce our proposed {\em Analyst}. Like the user model, Analyst is also an RL agent and care is needed not to cause confusion. Where for the user model, RL learns a policy that models human cognitive control knowledge, for the analyst, RL learns an policy for choosing experiments and inferring parameters.

\subsection{User model}

\begin{figure*}
  \includegraphics[width=0.6\textwidth]{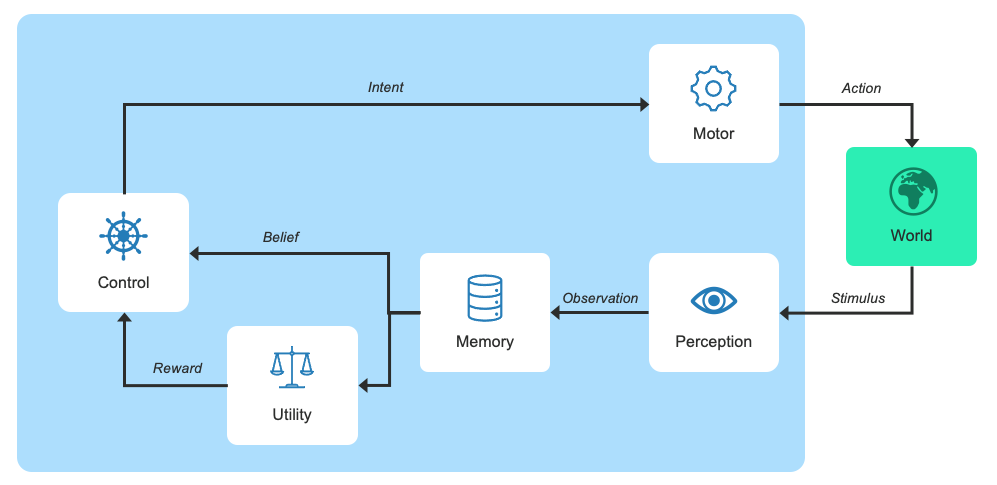}
  \caption{User model architecture. Human cognition is modelled as  five processes in interaction with a World. The Motor process models movement noise. Perception models increasing visual noise with eccentricity from the fovea. Memory integrates multiple observations using Bayesian inference and Utility generates a reward signal that captures the trade-off between factors such as speed and accuracy. }
  \Description{User model architecture. Human cognition is modelled as five processes in interaction with a World. The processes are control, motor, perception, memory and utility. A more detailed explanation is given in section 3.1.}
  \label{fig:user_model_architecture}
\end{figure*}

We extended a reinforcement learning model of gaze based target selection previously reported in \cite{chen2021adaptive}. The key assumption in the model is that the control of movement is computationally rational: that is, the saccade path and fixations  are determined by an attempt to optimise  some objective function (e.g. to minimise selection time) given the bounds imposed by the perceptual/motor system. The predicted eye movement strategies are therefore an adaptive consequence of the following  constraints: (1) target eccentricity, (2) target size, (3) oculomotor saccade noise, (4) a target detection threshold, and (5) location and target size estimation noise in peripheral vision. The interaction between the target size, target eccentricity, signal-dependent oculomotor noise, eccentricity-dependent estimation noise and size- and eccentricity-dependent target detection results in a multi-step gaze-based selection process. Two-step selections are typical in humans but under some circumstances either one-step (for large targets) or 3-plus-steps (very small targets) can be observed \cite{Meyer1988}. 

The user model architecture is illustrated in Figure \ref{fig:user_model_architecture}. The blue box contains processes (represented as white rectangles) that constitute a theory of human cognition in interaction with a `world'. Each trial begins with the controller choosing an `intent' -- a motor movement to an aim point (a location in the world). The chosen intent is implemented via a noisy `motor' process. The motor process results in an `action' which is the actual end point of the motor movement in the world. Additionally, the controller can perform a keypress that terminates the episode. Subsequently, a new stimulus is generated by the world which is the target location and width viewed from the new fixation. This stimulus is perceived by a foveated vision process that generates a noisy estimate of the target location and size (the `observation'). The observation provides evidence as to the location and size of the target.  This evidence  is optimally integrated in a Bayesian `memory' process which outputs a `belief'. The memory is observed by the `control' process and by a `utility' function. The utility function generates a reward signal that is used to train the controller. The controller's `intents' are thereby conditioned on the belief.
\footnote{Note, that the first action is selected before any observations are made. Therefore, while it is not conditioned on an observation, it may through training with the reward be informed by a prior expectation of the distribution of target locations.} The architecture implements an action-observation-reward cycle which repeats until the target is selected or the maximum step count for the episode is reached. Learning adjusts the mapping between the observation and the action so as to maximise the cumulative discounted rewards. The architecture can  be described formally as a POMDP.

\begin{itemize}
    \item \textbf{State space $S$}: At each time step $t$, the environment is in a state $s_t \in S$. A state represents a possible target position and width, and denotes as $s=(f_x,f_y,t_x,t_y,w)$ where $(f_x,f_y)$ is the fixation location and $(t_x,t_y)$ is the target location. For both, $x,y \in [-1,1]$, with $-1$ and $1$ being the edge of the display, and $w \in [0, 1]$.
    
    \item \textbf{Action space $A$ }: An action, $a_t$, is taken at each time step $t$. On each of these steps the controller decides where to attempt to fixate next (the aim point). An aim point is denoted as a coordinate $a=(a_x,a_y)$ where $a_x,a_y \in [-1,1]$. 
    
    \item \textbf{Reward function $r(s,a)$}: At each time step $t$, a reward is generated by a utility function that models the preference utilities of a user. We assumed that users can trade speed for accuracy. Faster speeds are accompanied by more errors. The reward at time $t$ is based on a linear gaze duration model $r(s_t,a_t) = -(\theta_{a} \times Amplitude(t) + \theta_{b})$, where the slope $\theta_{a}$ and intercept $\theta_b$ are parameters of the user. If the user performs a keypress, the episode is terminated and a value $r_{max} \times \theta_{preference}$ is added to the final reward if the target is fixated (gaze is within target radius), otherwise the reward is $-r_{max} \times \theta_{preference}$ if the target is not fixated (an error). 

    \footnote{Here the utility is conditioned directly on the state (cf. Figure \ref{fig:user_model_architecture}).}
    The parameter $\theta_{preference} \in [0,1]$ describes the speed-accuracy preference of the user. Additionally, if the maximum amount of steps is reached without a keypress, a termination penalty is added to the final reward.
    
    \item \textbf{Transition function $T(s_{t+1}|s_t,a_t)$}: The environment switches to a new state according to a stochastic transition function. The target location remains unchanged but the fixation location changes according to the outcome of the action aim point. 
    Aim points are corrupted by  noise. Therefore, $T(s_{t+1}|s_t,a_t) = N((f_x, f_y) | (a_x, a_y), \sigma_{ocular} (t))$. The oculomotor noise is linearly dependent on the saccade distance (the amplitude) $\sigma_{ocular}(t)= \rho _{ocular}\times Amplitude (t)$. 
    
    \item \textbf{Observation space $O$ and observation function $o=f(s,a)$}: After taking the action (i.e, saccade to and fixate at a new position on the display), a new observation is received, which is a function of state and action $o_t=f(s_t,a_t)$. The observation of the target position is dependent on the true target location and width (state), and the current fixation location (action). Specifically, 
    the spatial uncertainty of the target position (standard deviation) in peripheral vision is linearly dependent on the distance between the target and the current fixation position, i.e., eccentricity. We similarly assume a linear dependency between the uncertainty and the target size. Therefore, the perceived target position is $\tilde{t}_{x,t} \sim N(t_{x,t},\sigma_o (t))$, $\tilde{t}_{y,t} \sim N(t_{y,t},\sigma_o (t))$, where $\sigma_{o} (t)= \rho_{spatial} \times eccentricity (t) - \rho_{w} \times w + \rho_b$, $\rho_{spatial}$, $\rho_w$ and $\rho_b$ are parameters of the model. We also assume that the observed target width is corrupted by a Gaussian noise source $\tilde{w}_t \sim N(w_t, \sigma_w)$. Finally, a binary random variable $z_t \sim p(z|s_t,a_t)$, $z \in \{0,1\}$, indicates whether the user detects the target. Thus the full observation at time $t$ is the tuple $o_t = (\tilde{x}_t, \tilde{y}_t, \tilde{w}_t, a_t, z_t)$.
    
    \item \textbf{Discount rate $0 \leq \gamma\ < 1$). } The model receives a scalar reward at each time step, $r(s_t,a_t)$. The optimal strategy is the one that maximises the expected long-term sum of rewards: $E [\sum\limits_{t=0}^T \gamma^t r(s_t,a_t) ]$, given the constraints on the model defined above.
\end{itemize}

\subsubsection{Belief update}
If the target is detected by the user (i.e. if $z_t=1$ in the observation) the memory is updated by integrating the current belief $b_{t-1}$ and the new observation $o_t$ using Bayes rule (a Kalman filter) \cite{rao2010decision,dayan2008decision}. After taking an action (fixating at a location), the model receives noisy observations of the target location and width, which are sampled from a Gaussian distribution, $o_{t,k} \sim N(s_k,\sigma_{o,k}(t))$, with $k \in \{1,2,3\}$ (see Observation function above). We omit the index subscript $k$ for clarity. At the time step $1$, $b_1=o_1,\sigma_b^2(t=1)=\sigma_o^2(t=1)$. The belief update from $t$ to $t+1$ is shown in Equation (\ref{eq1}) below. 

\begin{equation} \label{eq1}
\begin{split}
b_{t+1} & = b_t + K_{t+1} [o_{t+1}-b_{t}]\\
{\sigma_b}^2 (t+1) &= {\sigma_b}^2 (t) -K_{t+1} {\sigma_b}^2 (t)\\
K(t_{t+1}) &=\frac{ {\sigma_b (t)}^2}{{\sigma_b(t)}^2+{\sigma_o(t+1)}^2} 
\end{split}
\end{equation}

\subsubsection{Training}
For each study, an ensemble model was trained, meaning that the parameters to be estimated were sampled from their prior distributions and given as additional inputs to the controller, thus distilling a population of user models covering the desired space of behaviours into a single set of neural network weights. The controller was implemented as a simple fully-connected feedforward neural network and optimised with the proximal policy optimisation (PPO) algorithm \cite{schulman2017proximal}. The POMDP was solved for the restricted class of policies that conforms to the task bounds outlined above.

\subsection{Analyst} 

\begin{figure*}
  \includegraphics[width=0.6\textwidth]{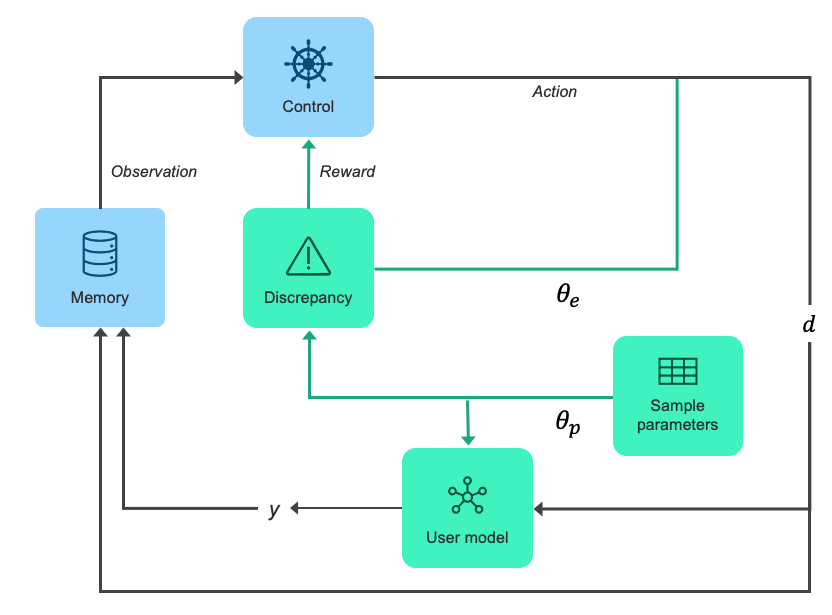}
  \caption{Analyst architecture. The problem for the reinforcement learner is to learn a sequential policy for the `control' process by acting in an environment that consists of a sequence of user models with stochastically sampled parameters. Actions consist of a $(d,\theta_e)$ pair where $d$ is a specification of the experiment and $\theta_e$ is an estimate of the model parameters. The `discrepancy' function calculates a reward by comparing the estimated parameters to the true parameters $\theta_p$. The `memory' stores the sequence of experiment designs $d$ and data $y$ that have been conducted for the latest sampled parameters. }
  \Description{Analyst architecture, consisting of control, memory, discrepancy, user model and parameter sampling from the prior. The details of the Analyst is given in section 3.2.}
  \label{fig:analyst-overview}
\end{figure*}

As was the case  for the user model, the environment of the Analyst is also formulated as a POMDP. The Analyst architecture is illustrated in Figure \ref{fig:analyst-overview}. The blue boxes form the reinforcement learning agent, and the green boxes represent the environment. The `user model' represents the simulator of a synthetic user. By sampling user parameters from a specified prior distribution, and conditioning the user model on these parameters, a range of different behaviours can be simulated. The user model is also conditioned, and its behaviour affected, by the design values produced by the controller. 
The parameterised user model generates data given a specific experimental design. The Analyst  tries various experimental designs to learn  about the user model and thereby estimate its parameters.

The `memory', which is considered as an internal process of the agent, stores the history of experimental designs and outcomes. The `control' process includes the policy function that effectively maps the contents of the memory to a probability distribution over actions. The action contains the design for the next experiment as well as estimations of the user parameters with data collected from past experiments. The 'discrepancy' unit is used during the training of the Analyst to produce the reward signal for the RL agent. More formally, the Analyst POMDP is specified as follows:

\begin{itemize}

\item \textbf{State space $S$}: At each time step $t$, the environment is in a state $s_t \in S$. A state $s_t=(d_t, y_t, \theta_p)$ represents the tuple consisting of the design value $d_t$ that was used to run the experiment at that time step, the experiment outcome $y_t$, and the user parameter vector $\theta_p$.

\item \textbf{Action space $A$ }: An action, $a_t$, is taken at each time step $t$. The action $a_t=(d_{t+1},\theta_e)$ tuple of the analyst includes designs $d_{t+1}$ for the next experiment and parameter predictions $\theta_e$ based on the information gathered so far.

\item \textbf{Reward function $r(s,a)$}: At each time step $t$, a reward is generated by a discrepancy function, which measures the similarity of predicted and true parameters of the user model as the negative L1 error \(r(s_t, a_t) = -|| \theta_p - \theta_e ||_1\). The reward is directly influenced by the ability of the analyst to estimate parameters, but it is crucially also indirectly influenced by the analyst's ability to design informative experiments.

\item \textbf{Transition function $T(s_{t+1}|s_t,a_t)$}: The environment switches to a new state according to the transition function. The user parameter vector $\theta_p$ is sampled from the prior $p(\theta)$ at the beginning of the episode, and remains fixed until the end of the episode. At each time step $t$, the user parameters $\theta_p$ and the design $d_{t+1}$ chosen by the analyst are used to run the simulator and produce an outcome \(y_{t+1} \sim p(y|\theta_p, d_{t+1})\), which gives the new state $s_{t+1} = (d_{t+1}, y_{t+1}, \theta_p)$.

\item \textbf{Observation space $O$ and observation function $o=f(s,a)$}: At each time step $t$, the state is passed through an observation function $o_t = f(s_t,a_t) = (d_t, \tilde{y}_t)$ before given to the analyst, where $\tilde{y}_t$ is a corrupted measurement of the true experiment outcome $y_t$. The user parameters $\theta_p$ are treated as a latent variable, they are included in the state but not the observation.

\end{itemize}

The partial observability of the environment motivates a policy that is conditioned on the full history of observations $o_{\leq t} = o_1o_2...o_{t}$. The analyst policy is a stochastic function that samples parameter predictions and designs for the next experiment conditioned on the observation history as $a_t \sim \pi^{analyst}(a_t | o_{\leq t})$. The objective of the analyst is to maximise the expectation of discounted return $E [\sum\limits_{t=0}^M \gamma^t r(s_t,a_t) ]$ where $M$ is the number of experiments performed for a specific user and $\gamma$ is the exponential discount rate. This objective is non-myopic since credit assignment for a particular experiment is performed based on the quality of all future parameter estimations.

\subsubsection{Policy network}

\begin{figure}
  \includegraphics[width=0.4\textwidth]{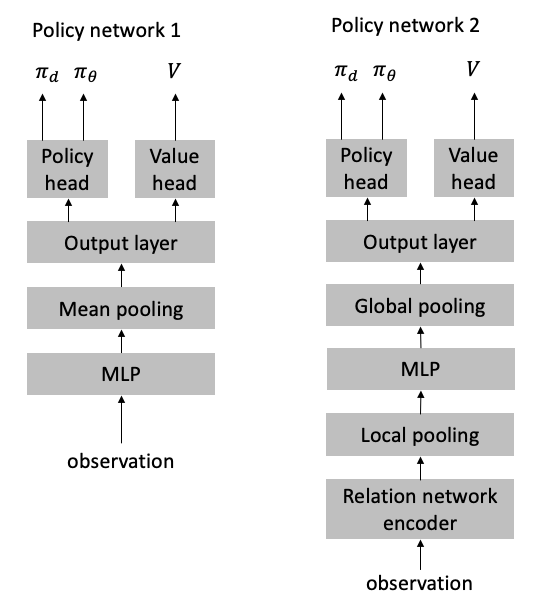}
  \caption{Two different policy network architectures.  Study 1 uses a simpler policy network, whereas  Studies 2 and 3 use a policy network architecture which implements relational reasoning.}
  \Description{Policy architecture. Study 1 uses a simpler policy network, whereas  Studies 2 and 3 use a policy network architecture which implements relational reasoning. A detailed description is given in chapter 3.2.1.}
  \label{fig:policy-architecture}
\end{figure}

The studies reported below use two different architectures for representing the policy network. 
In our first study, the observation includes information of the movement time and final fixation, target location and target width. In this case, the policy network implementation is a multilayer perceptron (MLP) network followed by mean pooling across experiments, output layer and heads for action distribution and value estimation (see figure \ref{fig:policy-architecture}, left). In studies 2 and 3, the observation includes eye movement data and information about the target location and width. In these studies, the policy network architecture uses an inductive bias to support relational inference \cite{santoro2017simple,battaglia2018relational} (see figure \ref{fig:policy-architecture}, right).

Considering the Studies 2 and 3 policy architecture, the output of the local pooling is an embedding of the observations over one episode,
\begin{equation}
e^l = \sum_{t=1}^T {f_{enc} (c^{target}, c_t^{fixation},c_{t+1}^{fixation})}
\end{equation}

where $f_{enc}$ denotes a relation network encoder (in our case an MLP network), conditioned on information of the target location $c^{target}$ and information of the locations of two subsequent fixations $c_t^{fixation}$ and $c_{t+1}^{fixation}$ at time steps $t$ and $t+1$. The output of the global pooling is an embedding over $M$ experiments

\begin{equation}
e^g = \sum_{i=1}^M{g(e_i^l)}
\end{equation}

where $M$ is the number of experiments and $g(.)$ is an MLP network. The embedding $e^g$ is fed to an fully connected output layer and to policy and value heads.

User models formulated as POMDPs, instead of closed form models, offer the possibility of capturing and simulating complicated and realistic human-like behaviour, but at the same time they raise some challenges. With such user models, we are often only able to  draw samples from the simulator, without the possibility of evaluating the likelihood or differentiating through the simulator. Because of this, new methods are needed that fulfill this requirement. In addition to the requirement for non-differentiable user-models, other desired properties are amortisation, non-myopic designs, and adaptation to the experiment outcomes. Table \ref{table_comparison} summarises the capabilities of various recent approaches for optimal experimental design. Our method is the most flexible as it is able to generate amortized, non-myopic and adaptive solutions in likelihood-free environments without the need to differentiate through the simulator.

\begin{table*}
\caption{Comparison of the capabilities of recent methods for optimal experimental design. * In MINEBED, a backup method is also described based on Bayesian optimisation, which does not require differentiable simulator.}
\begin{tabularx}{0.8\textwidth} { 
  | >{\raggedright\arraybackslash}X 
  | >{\centering\arraybackslash}X 
  | >{\centering\arraybackslash}X
  | >{\centering\arraybackslash}X 
  | >{\centering\arraybackslash}X 
  | >{\centering\arraybackslash}X | }
 \hline
 Algorithm & Amortized & differentiable simulator not needed & non-myopic & Adaptive & likelihood-free \\
 \hline
 DAD \cite{foster2021deep}  & \ding{52}  & \ding{52} & \ding{52} & \ding{52} &  \ding{55}  \\
\hline
 iDAD \cite{ivanova2021implicit}  & \ding{52}  & \ding{55}  & \ding{52} & \ding{52} & \ding{52}   \\
 \hline

  Blau et al \cite{blau2022optimizing}   &  \ding{52}  & \ding{52} & \ding{52} & \ding{52} & \ding{55}  \\
 \hline
  MINEBED \cite{kleinegesse2020bayesian}  &  \ding{55} & \ding{52}* & \ding{52} & \ding{52} &  \ding{52}  \\
 \hline
  Valentin et al \cite{valentin2021bayesian}  &  \ding{55} & \ding{52} & \ding{52} & \ding{52} & \ding{52}   \\
 \hline
  Our method  & \ding{52}  & \ding{52} & \ding{52} & \ding{52} & \ding{52}   \\
 \hline
\end{tabularx}

\label{table_comparison}
\end{table*}
\subsection{Training}

The policy network is trained by using both score-based and pathwise gradient estimators \cite{mohamed2020monte}. Since the design values are sampled from the action distribution, it uses a score-based gradient estimator implemented with PPO \cite{schulman2017proximal}. Since the parameter estimations can be directly optimised with a loss function, extra parameter updates can be conducted with the pathwise gradient estimations thereby reducing  variance.

The training method uses an exponential moving average (EMA) method in order to regularise the training process \cite{tarvainen2017mean}. The parameters of a separate EMA network are updated based on the parameters from the policy network, and the previous values of the EMA network:

\begin{equation}
\phi_t^{'}=\alpha\phi_{t-1}^{'}+(1-\alpha)\phi_t
\end{equation}

where $\phi_t^{'}$ is the EMA network parameter vector at time step $t$, $\phi_t$ is the parameter vector of the policy network, and $\alpha$ is the smoothing coefficient hyperparameter. The rewards for the policy network parameter updates are calculated by using parameter predictions from the EMA network, instead of the actual policy network. As a result, the policy parameter updates are less noisy and training is more efficient. 

The training was conducted by using the Stable Baselines 3 (SB3) library \cite{stable-baselines3} by implementing custom policy networks for the relation network, and by using the SB3 callback system for EMA implementation.

\section{Non-myopic and adaptivity demonstrations}
Our method produces amortised design strategies and parameter estimations in likelihood free settings without the need to differentiate through the simulator. As presented in Table \ref{table_comparison}, the existing methods are not applicable to our setting. To gain confidence for our method, its performance and range of applicability, we ran experiments on abstract tasks requiring non-myopic design strategies and adaptivity.
\subsection{Non-myopic demonstration}
\begin{figure*}
\includegraphics[width=0.9\textwidth]{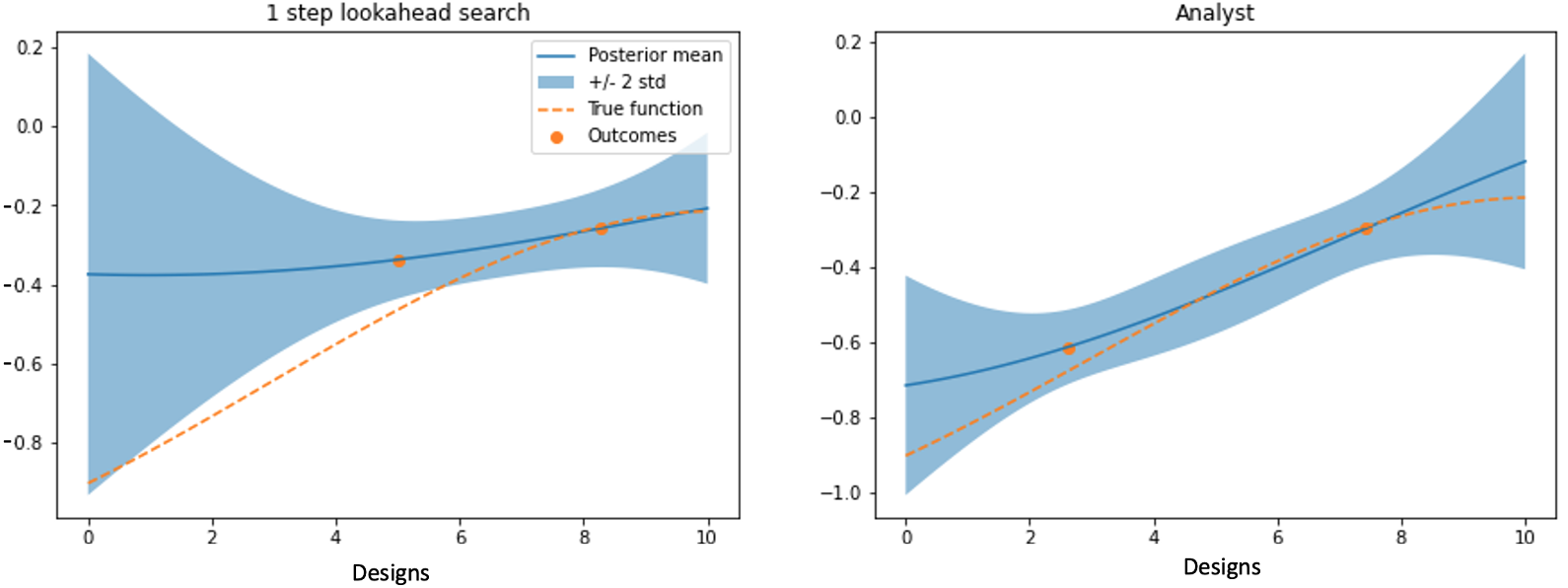}
  \caption{Examples of design selections by myopic and our non-myopic methods. Left panel: A myopic method has selected a design  to reduce variance with optimal one-step lookahead. Right panel: Our method has learned a non-myopic design strategy and can select more informative design values as the designs are chosen to more evenly  cover the whole space.}
  \Description{Examples of design selections by myopic and our non-myopic methods. The figures illustrate how the non-myopic method learns to select more informative designs that reduce the posterior uncertainty more efficiently.}
  \label{fig:nonmyopic1}
\end{figure*}

We demonstrate first the capability of our method to produce non-myopic design selection strategies with an abstract task. We use a 1-dimensional Gaussian process to sample functions with a specified kernel. Depending on the selected kernel, nearby points are correlated. As a baseline for comparison to our approach, we use a one step lookahead algorithm which minimises expected uncertainty reduction over the design space. This myopic algorithm will converge to non-optimal design strategy with two data points as the algorithm selects designs that are less uniformly distributed. In contrast, our non-myopic algorithm is able to take into account the total number of trials to be conducted and can design a whole series of experiments that lead to higher information gain overall.
The figure \ref{fig:nonmyopic1} illustrates examples of design selections of both algorithms. In the left panel, the  design is calculated with one-step lookahead so as to optimally reduce uncertainty. The mid-point design is chosen first in 'ignorance' of the fact that another experiment is to be conducted and as a consequence the overall distribution of experiments is not optimal. 
In the right panel, our non-myopic algorithm has successfully learned to choose designs that more evenly cover the design space, thereby gaining more information over the whole series of experiments and optimally reducing uncertainty overall.

The discrepancy function for the Analyst is the L2 distance between the true and estimated function.
We measured the quality of the designs with two metrics, namely the L2 distance between the estimated and ground truth functions, and the reduction of variance as calculated with the integrated mean-squared error (IMSE) method as used in Chen et al. 2019 \cite{chen2019adaptive}. As the myopic algorithm is analytically calculated to reduce variance, the latter metric offers a reliable comparison to the baseline.
The results are shown in table \ref{table_nonmyopic}, indicating a clear benefit of the non-myopic method.

\begin{table}[h!]
\caption{Results of the non-myopic experiment. The Analyst learns a non-myopic design strategy that outperforms the optimal myopic design strategy. The mean and standard error was computed over an evaluation batch with 100 functions sampled from the prior.}
\centering
\begin{tabular}{||c c c||} 
 \hline
 Metrics & mean & standard error \\ [0.5ex] 
 \hline\hline
 Analyst discrepancy  & 0.0812 & 0.0089 \\ 
 Baseline discrepancy & 0.1580 & 0.0204 \\
 Analyst IMSE & 0.3909 & 0.1194 \\
 Baseline IMSE & 0.7087 & 0.0 \\
 \hline
\end{tabular}
\label{table_nonmyopic}
\end{table}

\subsection{Adaptivity demonstration}
\begin{figure*}
\includegraphics[width=0.8\textwidth]{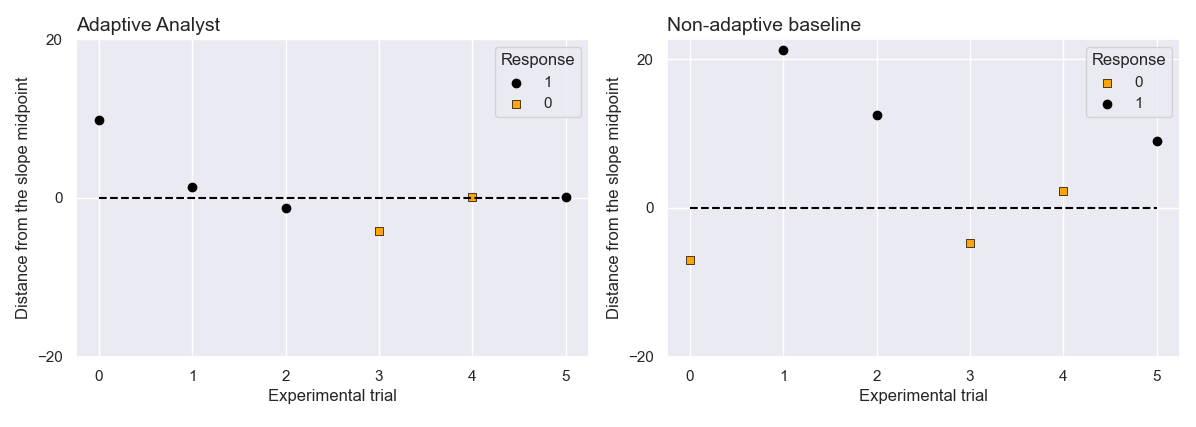}
  \caption{The Analyst learns to adapt to the data collected so far during the episode. Left panel: When the Analyst has information about the previous trials, it adapts to them and converges to the midpoint of the sigmoid function slope, which provides most information. Right panel: When information about the previous trial outcomes are not available, the Analyst learns a non-adaptive strategy in which the design values are sparsely distributed rather than focused where they are needed.}
  \Description{Illustration how the Analyst learns to adapt to the data collected so far during the episode. The figure illustrates how the adative version of the Analyst adapts to previous trials and converges to the midpoint of the sigmoid function slope.}
  \label{fig:adaptivity1}
\end{figure*}

The adaptivity is demonstrated in a setting where the task is to estimate a parameter that affects the positioning of a logistic sigmoid function on the x-axis. The data is generated by conducting Bernoulli trials where the probability $P(y_i = 1)$ is defined by a logistic function:
$$ P(y_i = 1|\theta, d) = \frac{1}{1+e^{-{(d+\theta)}}} $$
where $y_i$ is the outcome of the $i$:th trial in the experiment, $\theta$ is the parameter to be estimated and $d$ is the design value.
To train a non-adaptive baseline, we masked the outcomes from the Analyst until the last time step of the episode. In contrast, when the outcomes of the previous trials during the episode are available, the Analyst can generate an adaptive strategy where design choices are affected by the previous outcomes during the episode. To compare Analyst to the baseline, we use the MSE of the estimated parameter. The table \ref{table_adaptivity} indicates a clear benefit of the adaptive design strategy over the non-adaptive strategy. The adaptivity of the designs is illustrated in Figure \ref{fig:adaptivity1} which shows how the Analyst design selections converge on the mid-point of the slope in the sigmoid function.

\begin{table}[h!]
\caption{Results of the adaptivity demonstration. To provide a baseline, the Analyst trained without information of the previous outcomes during the episode learns a non-adaptive strategy. In contrast, when previous outcome information is available, the Analyst produces an adaptive strategy which beats both the non-adaptive baseline and a random baseline. The MSE mean and MSE standard error was computed over an evaluation batch with 10000 models sampled from the prior.}
\centering
\begin{tabular}{||c c c||} 
 \hline
 Method & MSE mean & MSE standard error \\ [0.5ex] 
 \hline
 Adaptive Analyst  & 2.018 & 0.034 \\
 Non-adaptive baseline & 6.265 & 0.085 \\
 Random design baseline & 14.434 & 0.234 \\
 \hline
\end{tabular}

\label{table_adaptivity}
\end{table}

\section{Study 1: Estimation of submovement noise from summary data}

Following Meyer's law we assume that people make a series of submovements to achieve a goal \cite{Meyer1988a}. In the first study, we apply our approach to a scenario where only summary statistics are provided to the analyst at the end of each episode. The summary statistics includes movement time from the beginning of the episode until the end of the episode, the target location, target width and the location of the final submovement. Although individual submovements are not observed, the goal is to infer the noise parameter affecting these movements. The movement time is calculated with the formula $mt = \sum_i^I({\theta_{a}x_i + \theta_{b}})$, where $I$ is the number of user steps within the episode, $x_i$ is the distance of the submovement during the current user step $i$, $\theta_{a}$ is the slope parameter, and $\theta_{b}$ is the intercept parameter of the movement time model. In Study 1, (though not in subsequent studies) both of these movement time parameters are fixed. The goal in Study 1 is to estimate the movement noise parameter.

\subsection{Results} 

\begin{figure*}
  \includegraphics[width=0.9\textwidth]{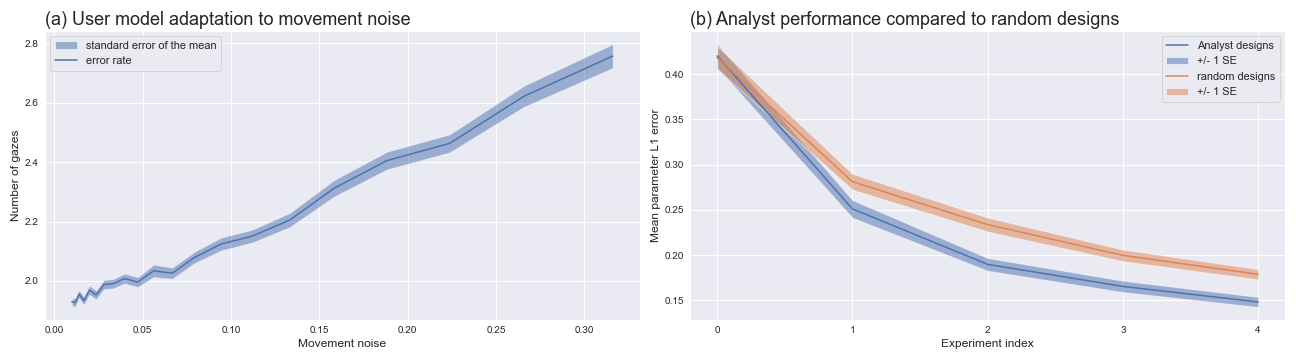}
  \caption{Study 1 results. Panel (a), The number of steps required by the user model to reach the target against increasing movement noise values. Panel (b), The Analyst makes better parameter estimations the more it makes experiments. The analyst also performs better across stages compared to a baseline that samples designs uniformly. In panel (a), the shaded area represents $\pm$ 1 standard error over an evaluation batch with 1000 user episodes. In panel (b), the shaded area represents $\pm$ 1 standard error over an evaluation batch with 1000 models sampled from the prior.}
  \Description{The figure shows how the user model adapts to the movement noise. In the left panel, it is shown how increasing the movement noise value increases the average number of gazes needed to reach the target. The right panel shows the improved performance against the random designs baseline.}
  \label{fig:study1_result2}
\end{figure*}

The larger the movement noise,
the further the actual movements can be from the intended aim points. As a result, with large noise values the user model requires more steps to reach the target. In order to verify that the user model has adapted to the various movement noise values, we plot the number of steps that the user model needs to reach the target on average. Figure \ref{fig:study1_result2}a shows that the fully trained user model require more steps to reach the target when the noise levels are higher. As the user model's behaviour is impacted by the parameter value, it should be possible to train the Analyst to infer the movement noise from behavioural observations.

\begin{figure*}
  \includegraphics[width=0.9\textwidth]{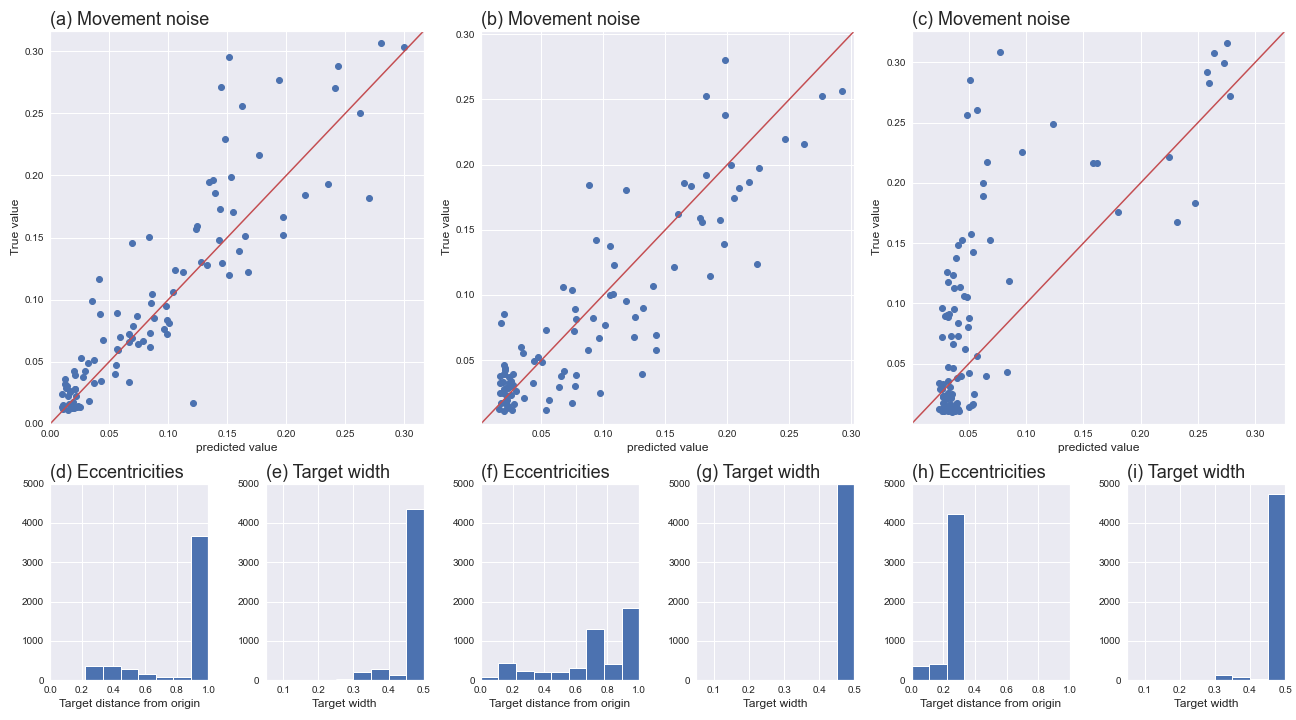}
  \caption{Study 1 results. Panels a-c illustrate the accuracy of the movement noise estimations when the user model has low (panel a), medium (panel b) or high (panel c) perceptual noise. Below the scatter plots, the corresponding design selections (target distance from origin and width) by the analyst are illustrated as histograms (panels d-i).}
  \Description{Study 1 results with scatter plots of the parameter estimation and the histograms of the design selections. The detailed analysis is in chapter 5.1}
  \label{fig:study1_result1}
\end{figure*}

We tested the ability of the Analyst to estimate parameters and the results are reported in Figure \ref{fig:study1_result1}. In each of the panel's (a), (b) and (c) the true value of the movement noise is plotted against the estimated value. A linear regression fit was performed to all three scatter plots ($a=0.99, b=0.91, R^2=0.80$ for panel (a), $a=0.83, b=0.01, R^2=0.77$ for panel (b), $a=1.00, b=0.03, R^2=0.54$ for panel (c)). The panels differ in the level of the perceptual noise, with low perceptual noise in panel (a), higher perceptual noise in panel (b) and highest perceptual noise in panel (c). Across all three panels it is clear that the Analyst is able to provide some level of estimate of the oculomotor noise, however, it is also clear that the estimates are much better when the perceptual noise is lower (panel (a)).  

Figure \ref{fig:study1_result1} also illustrates the distribution of experimental design choices made for each level of perceptual noise. Panels (d) and (e), which correspond to panel (a), show that for low perceptual noise, experimental designs tend toward higher eccentricities and large targets (though there is some variance).
However, as perceptual noise increases, the analyst is incentivised to select targets closer to the origin, in order to avoid corrupting the data with very noisy observations. This claim is supported by panels (f) and (h) which show more experiments with smaller eccentricities at higher perceptual noise values.

Finally, the quality of the design optimisation can be verified by training the analyst with random design values, and comparing the accuracy of the parameter estimations with an analyst trained to optimise design selections. The right panel of  Figure \ref{fig:study1_result2} is a plot of the error in the estimate against the experiment number. With zero experiments there is no data and the parameter estimate is simply the learned prior of the parameter distribution. After the data from each experiment are incorporated into a new parameter estimate, the error decreases, and it decreases more quickly with Analyst designed experiments. It demonstrates a clear improvement in the accuracy of the parameter estimations with optimised designs.

\section{Study 2: Inference of perceptual and motor noise from gaze movements}

In Study 2, we applied the Analyst to a user model of eye-movements.  Instead of using summary statistics as in Study 1, we allowed  Analyst to observe the gaze fixations of each step during the episode. We also extended the model with a target detection requirement. As a consequence, whether or not the user model observes the target is probabilistic. The probability of not observing the target increases when the target size is far away and target width is small. 

In this study we report the results of the Analyst inferring three parameters: oculomotor noise, perceptual noise and movement time intercept. The observations include fixations at each time step, duration of each gaze, information about the target location and target width. A key difference between Study 1 and Study 2 is that, where there is a fixed amount of data per experiment in Study 1, in Study 2, experiments that with more distance and smaller targets can generate more data. As we will see, this fact impacts the selected designs.

\subsection{Results} 

\begin{figure*}
  \includegraphics[width=0.9\textwidth]{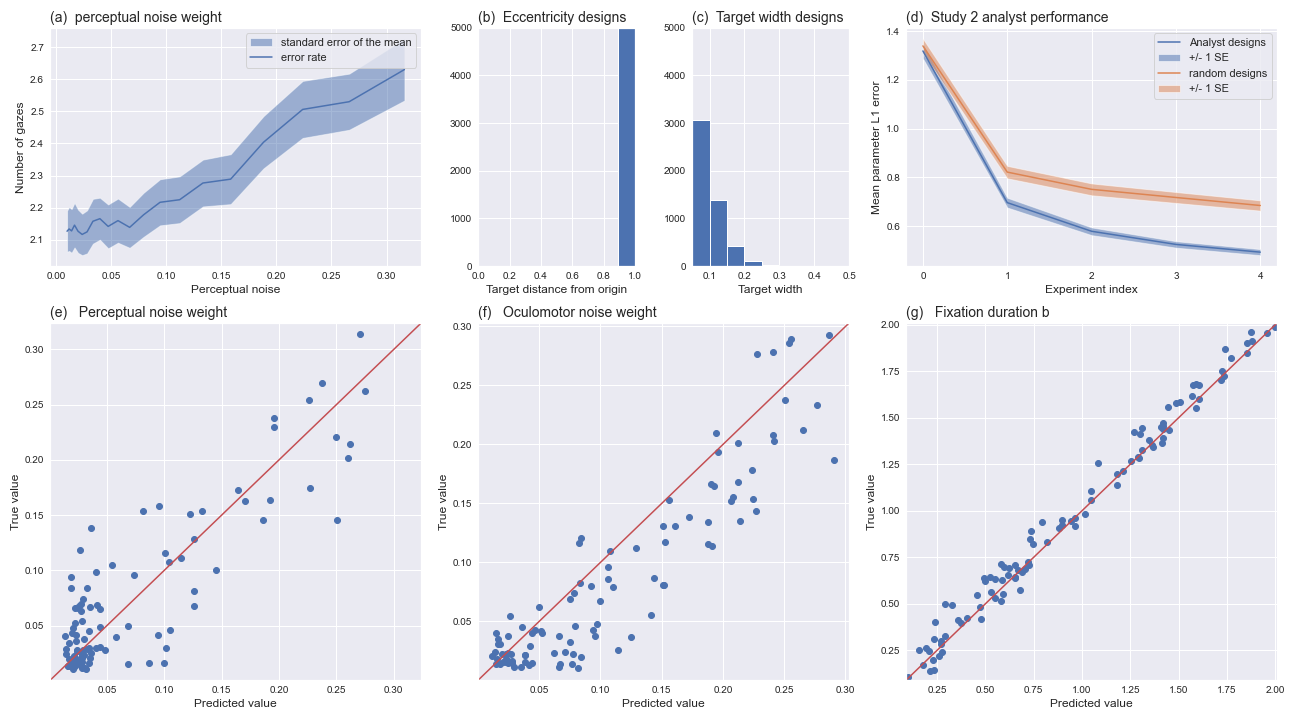}
  \caption{Results of  Study 2. Panel (a) shows the effect of the user model's perceptual noise parameter on number of steps taken. Panels (b) and (c) show the distribution of the experimental designs chosen by the Analyst. Panel (d) shows the the improvement in parameter estimates with Analyst designed experiments versus with random experiments. The panels on the lower row show the accuracy of the parameter estimations for perceptual noise (e), oculomotor noise (f), movement time interception parameter (g). In panel (a), the shaded area represents $\pm$ 1 standard error over an evaluation batch with 1000 user episodes. In panel (d), the shaded area represents $\pm$ 1 standard error over an evaluation batch with 1000 models sampled from the prior.}
  \Description{Study 2 results. The figure includes separate plots for user model adaptation to the perceptual noise weight, histograms of the selected design values, performance against random designs baseline and scatter plots of the parameter estimations. The detailed analysis is in chapter 6.1}
  \label{fig:study2_results}
\end{figure*}

As with Study 1, we first measured the effect of changes in the user model parameters on the behaviour. When considering oculomotor and perceptual noise values, increasing one of these noise values while keeping another fixed should cause the user model to require more gaze fixations to reach the target. This is clearly visible in the panel (a) in Figure \ref{fig:study2_results}, where the perceptual noise value versus required gaze steps to reach the target is plotted. The near linear increase in number of steps with increasing noise suggests that the parameter should be readily recoverable. The oculomotor noise is equivalent to movement noise in study 1, the effect of which is shown in Figure \ref{fig:study1_result2} of Study 1.

In Study 2, the goal of Analyst is to select the most informative experiments for inferring the oculomotor noise, perceptual noise and the intercept parameters for the movement time model. Analyst performance is illustrated in panels (e) to (g) of Figure \ref{fig:study2_results}. A linear regression was performed to assess the quality of the fits for each parameter ($a=0.85, b=0.01, R^2=0.76$ for panel (e), $a=0.87, b=-0.01, R^2=0.84$ for panel (f), $a=1.00, b=0.04, R^2=0.99$ for panel (g)). The selected design values are illustrated in panels (b) and (c) in Figure \ref{fig:study2_results}. In this case, the analyst has learned to design experiments with small targets and large eccentricities. 

Finally, the performance of the analyst is compared against an analyst trained by using random experimental designs. Panel (d) in Figure \ref{fig:study2_results} shows a clear benefit of the optimised designs across experiments.

In summary, the results of Study 2 extend Study 1 by showing that the analyst can choose experimental designs and accurately infer multiple parameters at the same time. In addition, it can do so when each experiment returns a sequence of data (fixation locations and movement times) and not just point values. Finally, estimating parameters with selected designs outperforms doing so with random designs.

\section{Study 3: Inference of Preferences }

In Study 3, we test to see whether Analyst can discover user model preferences.  Preferences are important to HCI as they capture personal and sometimes discretionary  aspects of how a person wants to interact with a computer. Preferences include preferences for music genre and/or movie directors, for example, but here we focus on speed-accuracy trade-offs. As we have said, the speed-accuracy trade-off is a significant determinant of how people choose to interact with computers \cite{zhai2004speed} and is readily detectable in behaviour. The speed-accuracy trade-off determines the error rate which is a key property of interaction. The faster users attempt to perform a pointing task then the more errors that they make. In Study 3 we tested the extent to which Analyst was capable of estimating these preference parameters. To do so, we simulated a task in which the user model is able to end an episode by pressing a key on a keyboard. The observation space is extended to include information about whether the user has pressed a key or not. 

\subsection{Results}
 
\begin{figure*}
  \includegraphics[width=0.9\textwidth]{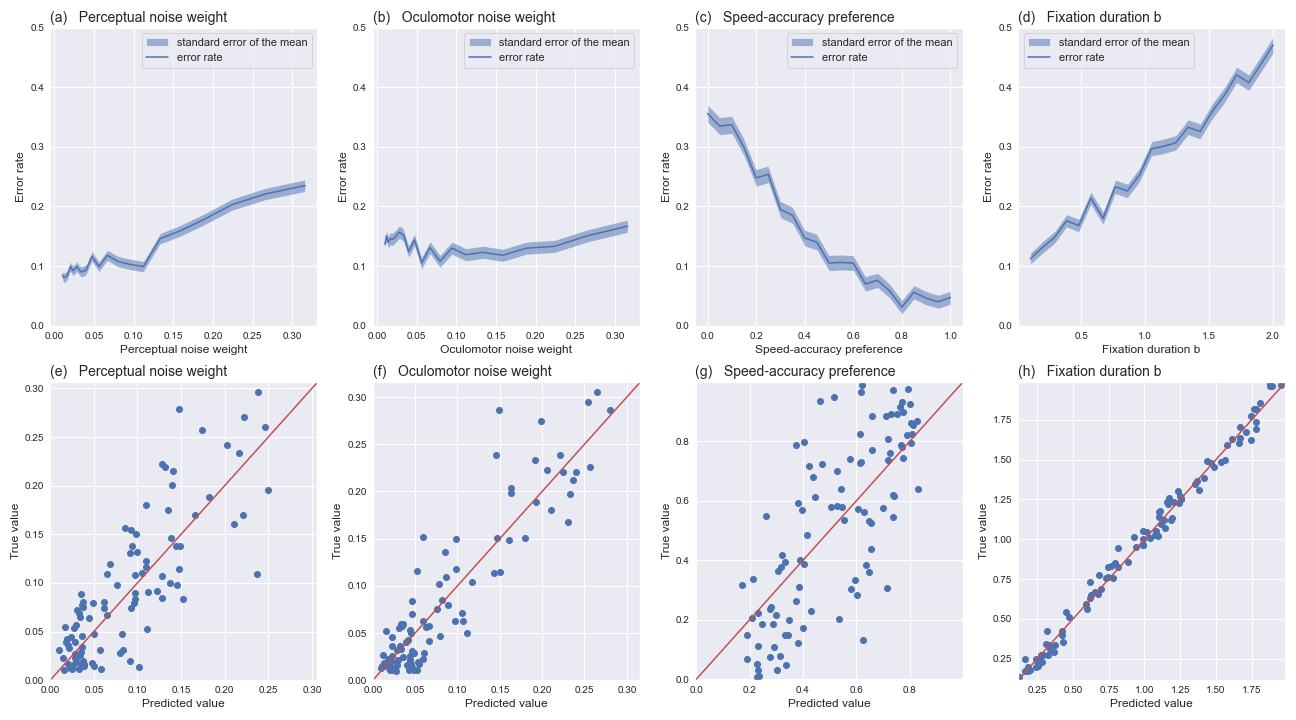}
  \caption{Results of Study 3. The upper row illustrates the adaptation of the user model to parameters for the perceptual noise (a), oculomotor noise (b), speed-accuracy preference (c) and movement time intercept (d). The lower row illustrates the corresponding accuracy of the parameter estimates. In panels (a) - (d)T, the shaded area represents $\pm$ 1 standard error over an evaluation batch with 1000 user episodes.}
  \Description{Study 3 results. The figure includes plots indicating how the user model adapts to the parameters and scatter plots of the parameter estimations. The detailed analysis is in chapter 7.1}
  \label{fig:study3_results1}
\end{figure*}

As with the previous studies, we first test whether the parameters, particularly the preference parameter, make an identifiable difference to the behaviour of the user model. In this case a useful metric for measuring the response of the parameters to the behaviour of the user agent is the error rate, which describes how often the user model ends the episode when the gaze is not in the true target, or the maximum number of gazes is reached without a keypress by the user. 
Figure \ref{fig:study3_results1}, panels (a) to (d) show the error rate for four of the model parameters. Panel (c) shows how the error rate is larger when the preference is biased towards speed, and approaches zero when the preference is biased towards accuracy.  

The lower row,  panels (e)-(h)  of  Figure \ref{fig:study3_results1} shows the capability of the analyst to simultaneously infer all four  parameters. As the Study 3 task is made more difficult by the increased number of parameters, the error of the estimates is greater when compared to Study 2. However, there is a good correlation between ground truth and parameter estimate for all four parameters. Having said that the worst of the four estimates is for the preference parameter which has a noticeable skew.

\begin{figure*}
  \includegraphics[width=0.9\textwidth]{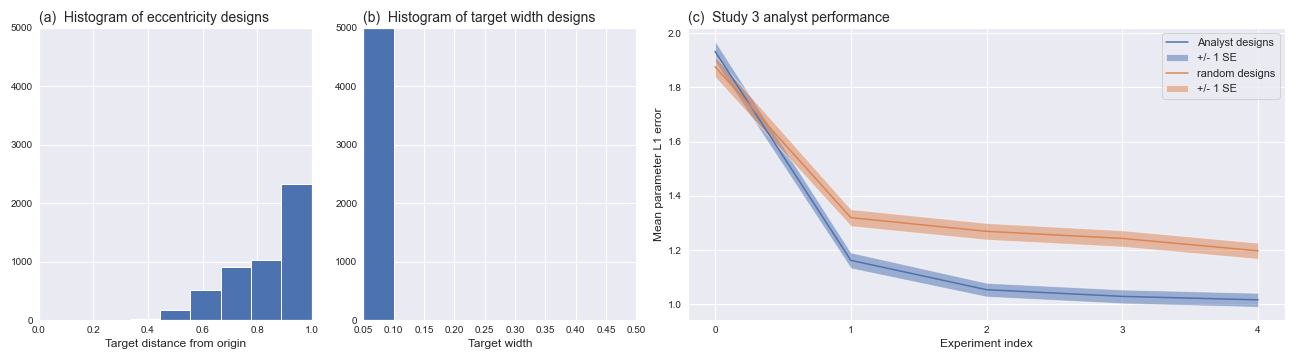}
  \caption{Results for Study 3.  Panels (a) and (b) show the histograms of Analyst selected experimental designs. Panel (c) shows the performance gain that follows from inferring parameters on the basis of the designs selected by the analyst compared to random designs. In panel (c), the shaded area represents $\pm$ 1 standard error over an evaluation batch with 1000 models sampled from the prior.}
  \Description{Study 3 results. The figure includes histograms of the design selections and the performance against random designs baseline. The detailed analysis is in chapter 7.1}
  \label{fig:study3_results2}
\end{figure*}

Figure \ref{fig:study3_results2} shows the Analyst's distribution of experimental design choices in two histograms. It always picks the smallest target (panel b) but shows a broader distribution of choice of distances. It is instructive to compare this distribution to that for Study 2.

In  Study 2, a good strategy for the analyst was to select a small target, as far away from the fovea as possible. In contrast, less extreme distances are selected in Study 3. One reason for this difference may be that because of the speed-accuracy trade-off there is a risk that the smallest most distant targets are not selected at all and therefore there is little evidence gathered to inform the speed-accuracy trade off preference parameter. Therefore, in Study 2 it makes sense to select experimental designs in which the user model stands some chance of high accuracy. Lastly, the Analyst's optimal experiments outperform an analysis conducted with data from random experimental designs (Figure \ref{fig:study3_results2} panel (c)).   

In summary, Study 3 demonstrates that  Analyst can simultaneously estimate multiple parameters, and importantly, it can estimate both capacity parameters (e.g. oculomotor and perceptual noise) and preference parameters (e.g. speed-accuracy trade off) -- from the same experiments. While there is a noticeable decrease in the quality of the parameter estimates with the increase in the number of parameters under consideration,  compelling correlations are still generated. As with the other two studies the analyst selected experimental designs outperformed the random designs.

\section{Discussion}

We have explored the properties of a new  method of user model parameter estimation  and shown that, for three progressively complex pointing tasks, it can make rapid estimates of parameter values for individual simulated users. It does so by learning a near-optimal policy for choosing the experiments that are most likely to generate informative data.

All three of the studies showed that the learned policies lead to more accurate parameter estimates than  random experiments. Parameters could be estimated both with data that summarised a sequence of user submovements (Study 1), as well as with data that included multiple steps in a single experimental observation (Studies 2 and 3) -- two important types of user data found in HCI. 
Further, in Study 1, 2 and 3, each of four successive designed experiment led to an improvement in the parameter estimates and a concomitant reduction in the prediction error. This improvement was more rapid for designed experiments than for random experiments. The estimated parameter values were highly correlated with the true value after only four experiments. 

While we have demonstrated the viability of the amortised approach for pointing tasks, more work is needed to verify that it generalises to other HCI tasks. While we believe that the approach is structured so as to provide design of experiments and inference of parameters for any complex simulation-based user models, further empirical work is needed. If generalisation is possible then, while very high computational costs are paid during training (between 3 and 9 hours of wall time on a laptop for the simulations reported above), the result is a very fast (milliseconds), data-lean (four observations) deployment. In other words, the approach trades the high cost of training an ensemble model and training the analyst, for a subsequent reduction in the time required to infer user model parameters. Post-training, not only can the best next experiment be determined in milliseconds but in addition the analyst can minimise the total number of experiments required to determine best fitting parameter values. However, further empirical work is required to see whether this promise is delivered for a broad range of HCI tasks.

Additional properties of our approach include that it provides adaptive, non-myopic designs (See the experiments in Section 4) and that it can be trained with arbitrary non-differentiable simulators with intractable likelihoods. As we see in the next Section, these properties compare well with other work in this area.

\subsubsection{Comparison to related approaches}

The work reported above was inspired and informed by recent work in both HCI and machine learning. Of particular importance was work on ensemble user models \cite{moon2022speeding, kwon2020inverse} and work on reinforcement learning based experimental design and inference \cite{blau2022optimizing}. Also of  importance is the work on Bayesian approaches to optimal experimental design \cite{myung2016model, cavagnaro2013optimal, foster2021deep, kleinegesse2020bayesian, kleinegesse2021gradient} and work on relation nets \cite{santoro2017simple} which was crucial to tractable reinforcement learning.

Bayesian experimental design has the clear potential advantage of mathematical rigour, as well as estimates of the posterior distribution of parameter values, rather than point estimates. The Bayesian framework, reviewed briefly above, tackles the problem of experimental design  by optimising the expected information gain e.g. how much more certain we will become about the values of the parameters we are fitting.  EIG is equivalent to maximising the Mutual Information (MI) between the parameters and data when performing the experiment design. However, EIG does not account for inaccuracies resulting from the amortisation in the parameter estimation. Thus, it is not clear that optimising EIG leads to good designs in our setting. Instead of optimising EIG, in the RL approach it is possible to directly optimise the designs for amortised parameter estimation through a joint objective. Also, estimating the MI in the conventional BOED framework is doubly intractable \cite{rainforth2018nesting,ivanova2021implicit}. Due to this computational complexity, estimating Bayes optimal designs is not feasible when doing experiments. This has led to approaches that amortise the cost to a pre-trained deep network by using a tractable approximation to the MI \cite{ivanova2021implicit,foster2021deep,blau2022optimizing}. Tractable computation of the MI objective for implicit models is further complicated by the fact that the likelihood function of the parameter is not known \cite{cranmer2020frontier,lintusaari2017fundamentals}. Ivanova et al. \cite{ivanova2021implicit} tackle this problem by introducing a separate critic network. Their approach however requires the use of a differentiable simulator. Using RL alleviates this need, as the score function gradient estimator directly calculates gradients from the specified reward.
   
\subsection{Future work}

The results reported above represent a preliminary investigation of the potential of RL-based experimental design and inference in HCI. They suggest a number of future studies. Perhaps most significantly, the effectiveness of the method must be tested with human participants. While the simulated participants used to test the approach above are sampled from the distribution of real users and previous studies have demonstrated the human-like behaviour of these simulations \cite{chen2021adaptive}, further work is needed. While human studies are beyond the scope of the current article, the software that we have built makes it very easy to deploy the Analyst learned policies in interactive software with an eye tracker. This software would -- without lag -- choose the best experimental design (target distance and width), observe a user's saccades and fixations, update model parameters, update the observation history and repeat.

The approach must also be tested on a broader range of tasks so as to empirically establish its generality. While we have formalised the method in terms that we believe to be fully general, in this paper, we have only tested it on abstract problems and pointing tasks. A broader range of HCI-related tasks would include menu-search tasks \cite{Chen2015emergence}, decision making tasks \cite{chen2017cognitive}, and biomechanical control tasks \cite{ikkala2022breathing}, to name but three.

In the future, user models with rapid parameter estimation could help  enhance interaction for each individual user. Mouse gain functions, text completion, icon sizes, colour pallet, etc. are almost always never tuned to an individual's preferences and capacities. Instead, interfaces provide settings by which the interaction can be `adapted' manually requiring the user to actively choose configurations in accordance with their beliefs about what is good for them. We believe that this process could be complemented with automatic personalisation methods based upon the RL-based Analyst reported above.

In addition, further work is needed on picking good hyperparameters for the Analyst. As the reported studies became more complex, training of the neural networks became more challenging and some amount of hyperparameter tuning was involved in generating the results reported above. Finding the very best possible performance with extensive hyperparameter search was not within the scope of the current article. Therefore, the performance for both optimised Analyst and Analyst using random designs could be improved. 

Lastly, further work is needed to explore the implications of RL-based amortised parameter estimation for a range of HCI-related problems. A/B testing, for example, can be enhanced by first fitting a user model and then selecting an interaction design accordingly. With amortised methods, it may be possible to do this in real time. Similarly, recommender systems and decision support systems might be enhanced by fitting simulation-based user models to user preferences and selecting recommendations with the aid of the fitted user model. 

In conclusion, we have demonstrated, in-silico, the potential utility of amortised experimental design and parameter estimation based on an RL algorithm that learns how to choose experiments for estimating user model parameters. By doing so, it can reduce the subsequent time cost of interaction with humans and the amount of human data.

\begin{acks}
  This work was supported by the Academy of Finland (Flagship programme: Finnish Center for Artificial Intelligence FCAI) and ELISE Networks of Excellence Centres (EU Horizon:2020 grant agreement 951847) and Bitville Oy. The authors want to thank the Probabilistic Machine Learning (PML) and the User Interfaces research groups at Aalto University for fruitful discussions and feedback. 
\end{acks}
\balance

\bibliographystyle{ACM-Reference-Format}
\bibliography{HowesA2021,chi2023}

\appendix

\section{Appendix: Architecture and hyperparameters details}
All simulations were run using PyTorch deep learning \cite{paszke2019pytorch} and Stable Baselines3 (SB3) Reinforcement Learning  \cite{stable-baselines3} libraries. As the focus on this paper is to show the capability of our method, we mainly use network architectures and hyperparameters that are typical for neural network training with only minor hyperparameter search.
\subsection{User model}
We use the same user model in Studies 1 and 2. The user model in Study 3 includes additional functionality for a key press.
The policy functions use a feature extractor network.
\begin{center}
\begin{tabular}[t]{ c c c c }
\small
Layer & description & dimension & activation \\ 
\hline
H1 & fully connected & 32 & ReLU \\  
H2 & fully connected & 64 & ReLU \\ 
H3 & fully connected & 128 & ReLU \\ 
H4 & fully connected & 128 & ReLU \\ 
\hline
\end{tabular}
\end{center}
The feature extractor is followed by the policy and value heads, which have identical architectures:
\begin{center}
\begin{tabular}[t]{ c c c c }
\small
Layer & description & dimension & activation \\ 
\hline
H1 & fully connected & 128 & ReLU \\  
H2 & fully connected & 64 & ReLU \\ 
\hline
\end{tabular}
\end{center}
SB3 produces Gaussian distributions for actions and value predictions, from where mean values are used as point estimates for actions and value predictions. We used the Proximal Policy Optimization (PPO) algorithm \cite{schulman2017proximal} to train the user model with Reinforcement Learning.  The most important PPO parameters are given in the table below.
\begin{center}
\begin{tabular}[t]{ c c }
\small
Parameter & value  \\ 
\hline
Training iterations & 3 000 000  \\ 
learning rate & 2e-4  \\  
gamma & 0.99  \\ 
clip range & 0.18  \\ 
entropy coefficient & 0.001  \\ 
maximum gradient norm & 0.55  \\ 
\hline
\end{tabular}
\end{center}

\subsection{Analyst model}
As described in the body text, the Analyst uses a relation network with a mean pooling MLP network. The Analyst has the following layers:
\begin{center}
\begin{tabular}[t]{ c c c c }
\small
Layer & description & dimension & activation \\ 
\hline
H1 & fully connected & 32 & ReLU \\  
H2 & fully connected & 64 & ReLU \\ 
H3 & fully connected & 128 & ReLU \\ 
H4 & fully connected & 256 & ReLU \\ 
H5 & - & pooling layer & no activation \\
H6 & fully connected & 256 & ReLU \\
H7 & fully connected & 64 & RelU \\
\hline
\end{tabular}
\end{center}
\hfill \break
The feature extraction network is followed by identical policy and value networks with the following dimensions:
\begin{center}
\begin{tabular}[t]{ c c c c }
\small
Layer & description & dimension & activation \\ 
\hline
H1 & fully connected & 64 & ReLU \\  
H2 & fully connected & 64 & ReLU \\ 
\hline
\end{tabular}
\end{center}
\hfill \break
As with the user model, SB3 produces Gaussian distributions for actions and value predictions, from where mean values are used as point estimates for actions and value predictions. Similarly, the Analyst is trained with the PPO algorithm. The most important PPO parameters are given in the table below.
\begin{center}
\begin{tabular}{ c c }
\small
Parameter & value  \\ 
\hline
Training iterations & 4 000 000  \\ 
exponentially decaying learning rate & start 5e-5  \\  
gamma & 0.99  \\ 
clip range & 0.10  \\ 
entropy coefficient & 0.01  \\ 
maximum gradient norm & 0.55  \\ 
\hline
\end{tabular}
\end{center}

\subsection{Non-myopic demonstration}
The Non-myopic demonstration was trained using the Soft Actor Critic (SAC) algorithm with SB3 default hyperparameters. The Analyst network is simplified so that the relation network is replaced by a multilayer perceptron, therefore data point pairwise calculations (defined in equations (2) and (3)) are not used. The policy and Q networks have 2 layers, each with 128 hidden units and ReLU activations.

\subsection{Adaptation demonstration}
The Adaptation demonstration uses PPO with SB3 default settings. The EMA network  (described in Section 3.3) is not used.
The feature extractor network consists of 4 Layers followed by ReLU activations. The dimensions of the layers are 256 for layer H1, 256 for layer H2, 128 for layer H3 and 256 for layer H4. This is followed by a mean pooling layer H5 and a layer H6 of width 256 followed by ReLU activations. The policy and value heads are identical with two layers with a width of 256 units followed by ReLU activations.

\end{document}